\documentclass[12pt]{article}

\usepackage{mathptmx} 

\usepackage{amssymb}
\usepackage{amsmath}

\usepackage[]{graphicx}

\usepackage[letterpaper,margin=1in]{geometry}

\renewenvironment{abstract}
	{\quotation}
	{\endquotation}

\date{}

\makeatletter
\renewcommand{\fnum@figure}{\textbf{Figure \thefigure}}
\renewcommand{\fnum@table}{\textbf{Table \thetable}}
\makeatother

\usepackage[numbers,sort&compress]{natbib}
\bibpunct{(}{)}{;}{n}{}{,} 

\usepackage{url}

\usepackage[utf8]{inputenc}
\usepackage[T1]{fontenc}

\def\scititle{
	Natural Language-Driven Global Mapping of Martian Landforms
}
\title{\bfseries \scititle}

\author{
	Yiran Wang$^{1\dagger}$, Shuoyuan Wang$^{1\dagger}$,
	Zhaoran Wei$^{1}$, Jiannan Zhao$^{2}$,\\
	Zhonghua Yao$^{3}$, Zejian Xie$^{1}$, Songxin
	Zhang$^{1}$, Jun Huang$^{4}$,\\
	Bingyi Jing$^{5}$, Hongxin Wei$^{1\ast}$
	\\
	\small$^{1}$School of Science, Southern University of Science and Technology, Shenzhen, China.\\
	\small$^{2}$China University of Geosciences, Wuhan, China.\\
	\small$^{3}$University of Hong Kong, Hong Kong SAR, China.\\
	\small$^{4}$Hubei Key Laboratory of Planetary Geology, China University of Geosciences, Wuhan, China.\\
	\small$^{5}$The Chinese University of Hong Kong, Shenzhen, China.\\
	\\
	\small$^\ast$Corresponding author. Email: weihx@sustech.edu.cn\\
	\small$^\dagger$These authors contributed equally to this work.
}

\begin{document}

\maketitle

\begin{abstract} \bfseries
Planetary surfaces are typically analyzed using high-level semantic concepts in natural language, yet vast orbital image archives remain organized at the pixel level. This mismatch limits scalable, open-ended exploration of planetary surfaces. 
Here we present MarScope, a planetary-scale vision-language framework enabling natural language-driven, label-free mapping of Martian landforms. MarScope aligns planetary images and text in a shared semantic space, trained on over 200,000 curated image-text pairs. This framework transforms global geomorphic mapping on Mars by replacing pre-defined classifications with flexible semantic retrieval, enabling arbitrary user queries across the entire planet in \textasciitilde5 seconds with F1 scores up to 0.978. Applications further show that it extends beyond morphological classification to facilitate process-oriented analysis and similarity-based geomorphological mapping at a planetary scale. MarScope establishes a new paradigm where natural language serves as a direct interface for scientific discovery over massive geospatial datasets.
\end{abstract}

\section*{Introduction}
\noindent
Planetary science increasingly depends on the interpretation of vast and
rapidly expanding remote-sensing datasets to reconstruct surface
processes and environmental evolution. Scientists analyze planetary
landforms using high-level geomorphological concepts, such as
aeolian \cite{sullivan2005aeolian, liu2023martian}, volcanic \cite{werner2009global, robbins2011volcanic},
fluvial \cite{baker2006geomorphological, fassett2008timing}, glacial \cite{forget2006formation}, and impact
processes \cite{head2010global}, expressed in natural language. In
contrast, planetary image archives are organized primarily by pixels,
rather than by semantic geological meaning. This mismatch between human
conceptual reasoning and data organization fundamentally limits
scalable, open-ended exploration of planetary surfaces.

To address this challenge, researchers have focused on generating global
geomorphological maps using either manual interpretation or
task-specific supervised learning approaches. Manual mapping remains the
gold standard for accuracy but is inherently slow, labor-intensive, and
difficult to scale beyond a limited number of landform types. Supervised
learning approaches can accelerate mapping for specific morphologies,
yet they require large, carefully labeled training datasets and
typically generalize poorly across heterogeneous landforms. As orbital
image archives continue to expand in both resolution and
volume \cite{malin2007context, mcewen2007mars}, these limitations increasingly constrain
discovery, leaving planetary science ``data-rich but
interpretation-limited''.

Recent advances in multimodal representation
learning \cite{radford2021learning} offer a promising pathway toward overcoming
these limitations. Contrastive vision language models trained on paired
images and text can learn high-level semantic representations that link
visual patterns with linguistic concepts, enabling zero-shot and
label-free retrieval using natural language prompts. Applied to
planetary data, such models have the potential to move beyond
pixel-level classification toward semantic representations that capture
not only morphological similarity but also relationships related to
surface processes and formation mechanisms, as encoded in language. Yet
this opportunity remains largely unrealized: there is still no
planetary-scale multimodal system for fast, label-free retrieval of
diverse landforms, nor any framework that delivers comprehensive,
quantitative, cross-comparable maps of planetary geomorphology.

\section*{Results}

\begin{figure}[htbp]
\centering
\includegraphics[width=0.95\linewidth]{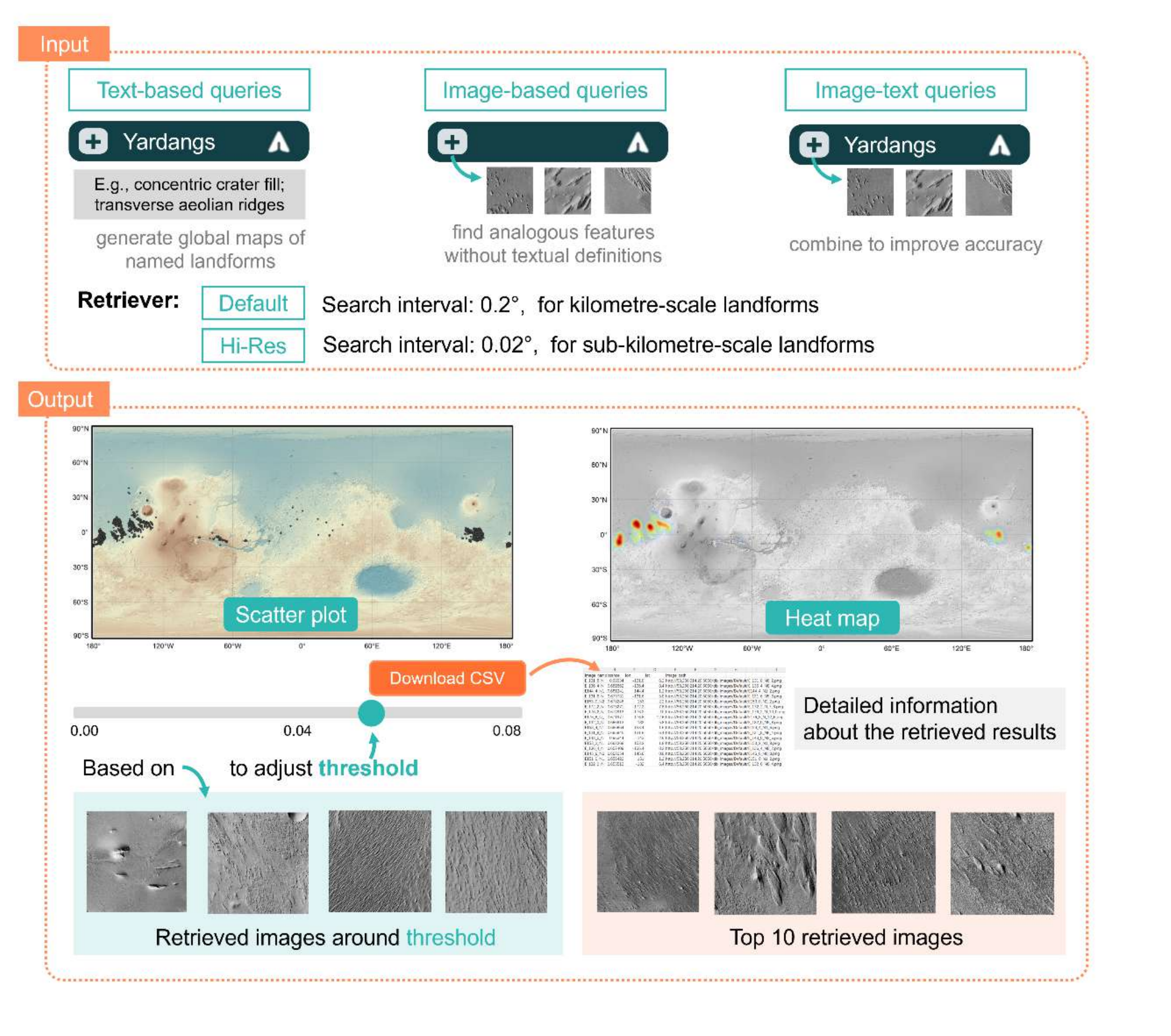}
\caption{\textbf{Workflow of the MarScope platform.} MarScope
enables three query modes---text-based, image-based, and image-text---to
retrieve Martian landforms. The system outputs global distribution maps,
heat maps, detailed records, and curated image sets, supporting both
mapping and dataset construction.}
\label{fig:1}
\end{figure}

\paragraph*{Near instantaneous planetary-scale retrieval of Martian landforms.}
Here we introduce MarScope (\url{http://marscope.site/}), a planetary-scale
visual-language platform that enables near-instantaneous
(\textasciitilde5 s) retrieval and global mapping of Martian landforms.
Trained on more than 200,000 curated planetary image--text pairs,
MarScope embeds surface imagery and language into a shared
vision--language space, allowing text prompts, representative images, or
their combination to serve as flexible search queries (Figure \ref{fig:1}). This
semantic framework redefines global geomorphological exploration as a
label-free retrieval task, enabling the identification of common
landforms by text such as name or formation process, as well as
previously unmapped features through images.

At its core, MarScope uses a contrastive vision--language
encoder \cite{radford2021learning} that learns high-dimensional embeddings
capturing diagnostic aspects of surface morphology. By representing
images and text within a joint semantic space, the platform allows
researchers to search, compare, and classify Martian surface features
flexibly and at scale. This shift from pixel-based imaging to structured
semantic understanding enables global mapping of any feature at any
time.

To accommodate diverse scientific use cases, MarScope supports three
complementary query modes. Text-based queries generate global
distribution maps of specified landforms or their formation process (for
example, yardangs, concentric crater fills, or shallow ground ice),
revealing spatial patterns that would otherwise require extensive manual
work. Image-based queries retrieve morphologically similar examples
without predefined labels, facilitating the discovery of rare or
understudied landforms such as doublet craters. Image-text multimodal
queries combine visual exemplars with textual constraints to refine
results for complex or heterogeneous morphologies.

The platform operates on a global mosaic of CTX
data \cite{malin2007context, dickson2018global, dickson2023release} subdivided into overlapping tiles at two
spatial resolutions: a 0.2° default mode optimized for kilometre-scale
features and a 0.02° high-resolution (Hi-Res) mode targeting
sub-kilometre landforms. Each tile is encoded once into the shared
embedding space, stored in a similarity-search index, and collectively
compresses the CTX dataset by a factor of \textasciitilde160. Incoming
queries---text, image, or image-text---are compared against all tiles
using cosine similarity, and the highest-scoring matches are projected
back onto a global map and visualized as point distributions or heatmaps
to show spatial extent and density.

In addition to visualization, MarScope allows direct download of matched
imagery, providing ready-to-use training samples for downstream
classification and recognition tasks. This greatly lowers the barrier to
dataset construction and accelerates the development of specialized AI
models.

\begin{figure}[htbp]
\centering
\includegraphics[width=0.9\linewidth]{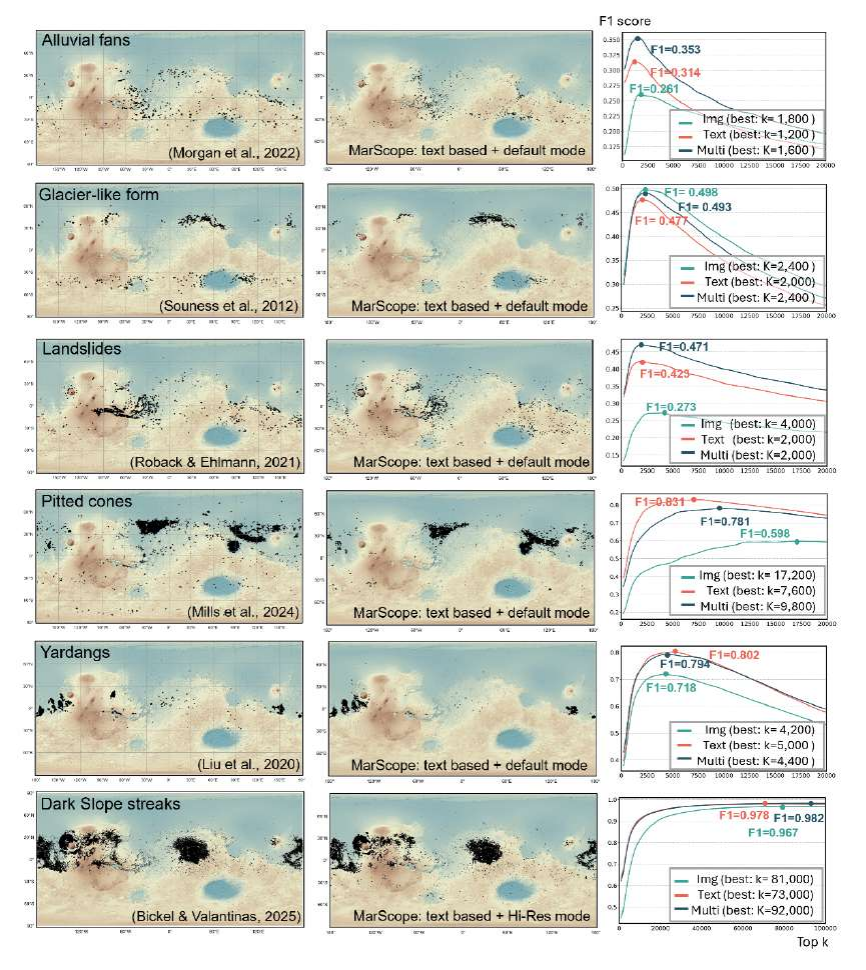}
\caption{\textbf{Validation of MarScope outputs.} Left column: Published
global distributions for six representative Martian landforms: alluvial
fans \cite{morgan2022global}, glacier-like forms \cite{souness2012inventory}, landslides \cite{roback2021controls}, pitted cones \cite{mills2024global}, yardangs \cite{liu2020mapping},
and dark slope streaks \cite{bickel2025streaks}. Middle column: corresponding MarScope
retrievals generated using text-mode queries, with the Hi-Res mode
applied to dark slope streaks and the default mode used for the other
five landforms. Right column: dynamic F1 scores as a function of the
top-K retrieved tiles for image (green), text (red), and multi-modal
(blue) query modes. Peak F1 values and the corresponding K are shown for
each landform type.}
\label{fig:2}
\end{figure}

\paragraph*{Retrieval performance evaluated against global geomorphological datasets.}
We evaluated MarScope's ability to recover known geomorphological
patterns by comparing its outputs with six published global catalogues
representing diverse surface processes: alluvial
fans \cite{morgan2022global}, glacier-like forms \cite{souness2012inventory},
landslides \cite{roback2021controls}, pitted cones \cite{mills2024global},
yardangs \cite{liu2020mapping} and dark slope streak \cite{bickel2025streaks}.
The first five classes were assessed using the default search mode,
whereas dark slope streaks---owing to their small spatial scale---were
evaluated using the high-resolution mode. For each landform type, we
tested three retrieval strategies: text queries (landform name only),
image queries (six representative examples), and multimodal queries
(combining text and images). Published catalogues and text-mode outputs
are shown in Figure \ref{fig:2}.

Retrieval fidelity was quantified using dynamic F1 scores as a function
of the top-K retrieved tiles (Figure \ref{fig:2}, right; computation described in
Methods). No single query mode performs best across all landforms,
reflecting differences in morphological distinctiveness and
representation within the training corpus. Pitted cones and yardangs
achieve their highest F1 scores with text queries, indicating that their
diagnostic morphology is well captured by semantic labels. In contrast,
alluvial fans and landslides benefit from multimodal queries, where
additional visual constraints improve performance. Although image-only
queries generally underperform relative to text and multimodal
approaches, they still retrieve relevant matches despite relying on only
six exemplars, highlighting the robustness of MarScope's visual
embedding space.

Overall, retrieval performance varies systematically with landform type,
and the ability to switch flexibly among text, image, and multimodal
queries is essential for optimizing results. Combined with the close
spatial correspondence between MarScope outputs and published
catalogues, these tests demonstrate that MarScope provides a reliable
and generalizable framework for rapid global reconstruction of Martian
landforms.

\begin{figure}[htbp]
\centering
\includegraphics[width=0.9\linewidth]{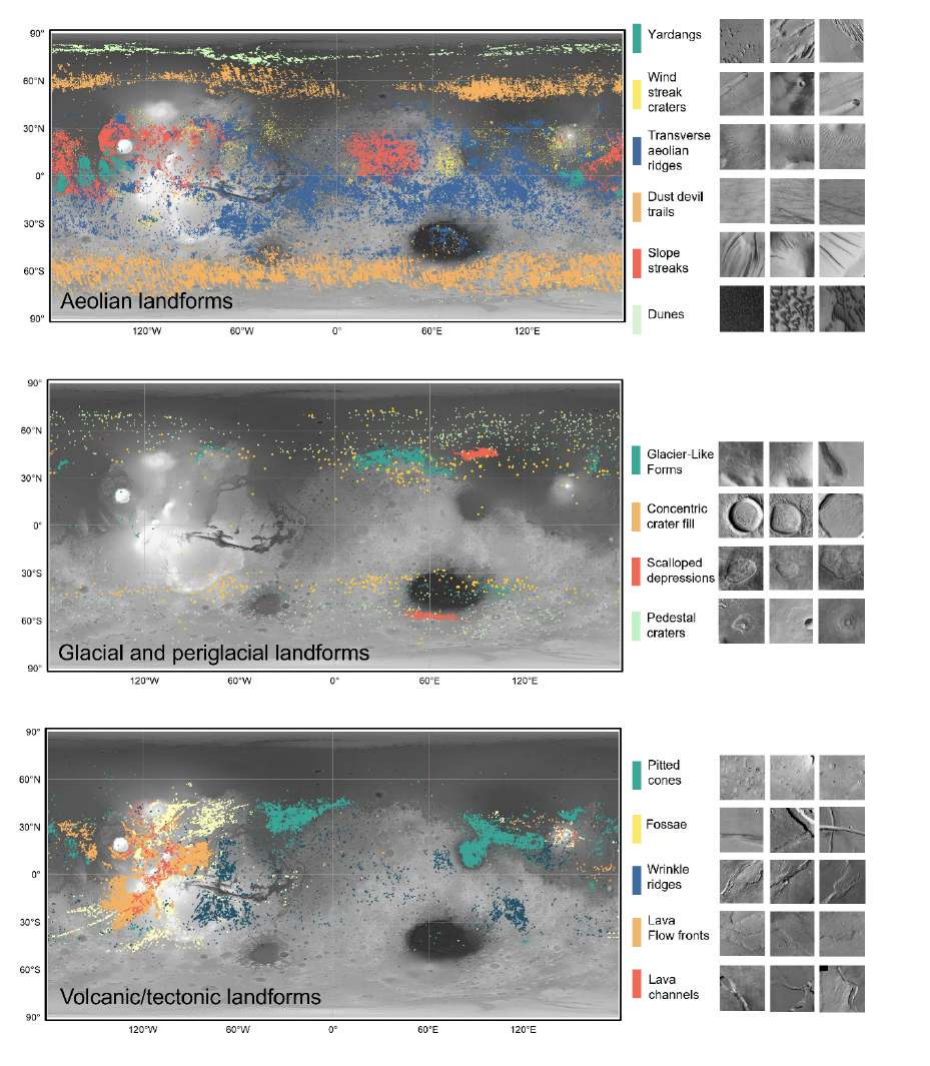}
\caption{\textbf{Landform distributions retrieved by MarScope.}
Global maps of aeolian (top), glacial and periglacial (middle), and
volcanic/tectonic (bottom) landforms on Mars. Colors indicate landform
classes, with representative CTX patches on the right showing
characteristic morphologies.}
\label{fig:3}
\end{figure}

\paragraph*{Global mapping of diverse Martian landforms.}
To enable planet-scale geomorphological analysis, MarScope was used to
generate global mapping of key Martian landforms across aeolian,
glacial-periglacial, volcanic, and tectonic systems, providing a unified
framework for addressing broad surface-process questions (Figure \ref{fig:3}).

The aeolian system of Mars is not a uniform global sand-transport field
but shows pronounced latitudinal partitioning. At the poles, extensive
sand dunes are actively reworked by the seasonal CO$_2$ frost cycle,
driving pulsed sediment transport \cite{portyankina2013observations, hansen2011seasonal}. Near
\textasciitilde60° N and \textasciitilde60° S, dust devil
trails \cite{daubar2025global} form high-activity belts that record strong
summertime convection. At lower latitudes (0--40°), longer-term aeolian
processes dominate: transverse aeolian ridges (TARs), dated to several
million years \cite{liu2023martian, berman2011transverse, reiss2004absolute, gou2022transverse, lu2022aeolian, berman2018high, kerber2012progression}, are widespread across low to
mid-latitudes, whereas yardangs cluster near the equator as relics of
ancient, high-energy wind erosion. Slope streaks \cite{bickel2025streaks, bickel2025dust}
and wind streaks \cite{sullivan2005aeolian, veverka1981wind}, concentrated in the northern
hemisphere, represent active surface disturbances and reflect
hemispheric asymmetry arising from topographic and dust-source
contrasts.

Glacial and periglacial landforms delineate a latitudinally zoned and
vertically stratified cryosphere. Pedestal craters (45°--70° N/S) record
the removal of metre- to tens-of-metres-thick ice-rich
mantles \cite{kadish2009latitude, kadish2010pedestal}. Glacier-like forms
(GLFs) \cite{hubbard2014glacier} and concentric crater fills
(CCFs) \cite{levy2010concentric} dominate the midlatitudes, preserving stable
ice--dust mixtures near the surface. In contrast, scalloped depressions
indicate shallow (\textless{} 20 m) ground-ice enrichment, particularly
concentrated in northern Utopia Planitia and southern Hellas, where
permafrost remains active today \cite{haltigin2014co, sejourne2011scalloped}.

Volcanic and tectonic landforms display strong spatial clustering and
coupling. The Tharsis and Elysium provinces host the most extensive
volcanic activity, characterized by fossae \cite{plescia1991graben, borraccini2005crustal}, lava
flows \cite{theilig1986lava, bleacher2007olympus}, and lava channels \cite{williams2005erosion}
arranged in radial and concentric patterns around major edifices. In
contrast, wrinkle ridges \cite{ruj2021global} are globally distributed,
recording compressional history and thermal evolution of Mars. Pitted
cones, commonly interpreted as products of magma--volatile
interactions \cite{mills2024global, chen2024global}, occur independently of major
volcanic centers but are concentrated in ancient sedimentary plains such
as Utopia and Isidis, reflecting localized interactions between
magmatism and volatile-rich substrates \cite{wang2023explosive, mcgowan2011utopia}.

Together, these results demonstrate that MarScope reframes
geomorphological mapping from a predefined classification task into an
open-ended semantic retrieval problem, enabling coherent, cross system
analysis of Martian surface processes at planetary scale.

\paragraph*{Process-oriented mapping through semantic retrieval.}
In addition to retrieving landforms by name, MarScope enables mapping of
geomorphological features through process-based queries, allowing
landforms to be explored according to their inferred formation process
mechanisms rather than predefined shapes or labels. This representation
allows landforms to be explored according to their inferred formation
mechanisms, rather than predefined shapes or fixed taxonomies,
addressing a long-standing limitation of traditional geomorphological
mapping.

\begin{figure}[htbp]
\centering
\includegraphics[width=0.92\linewidth]{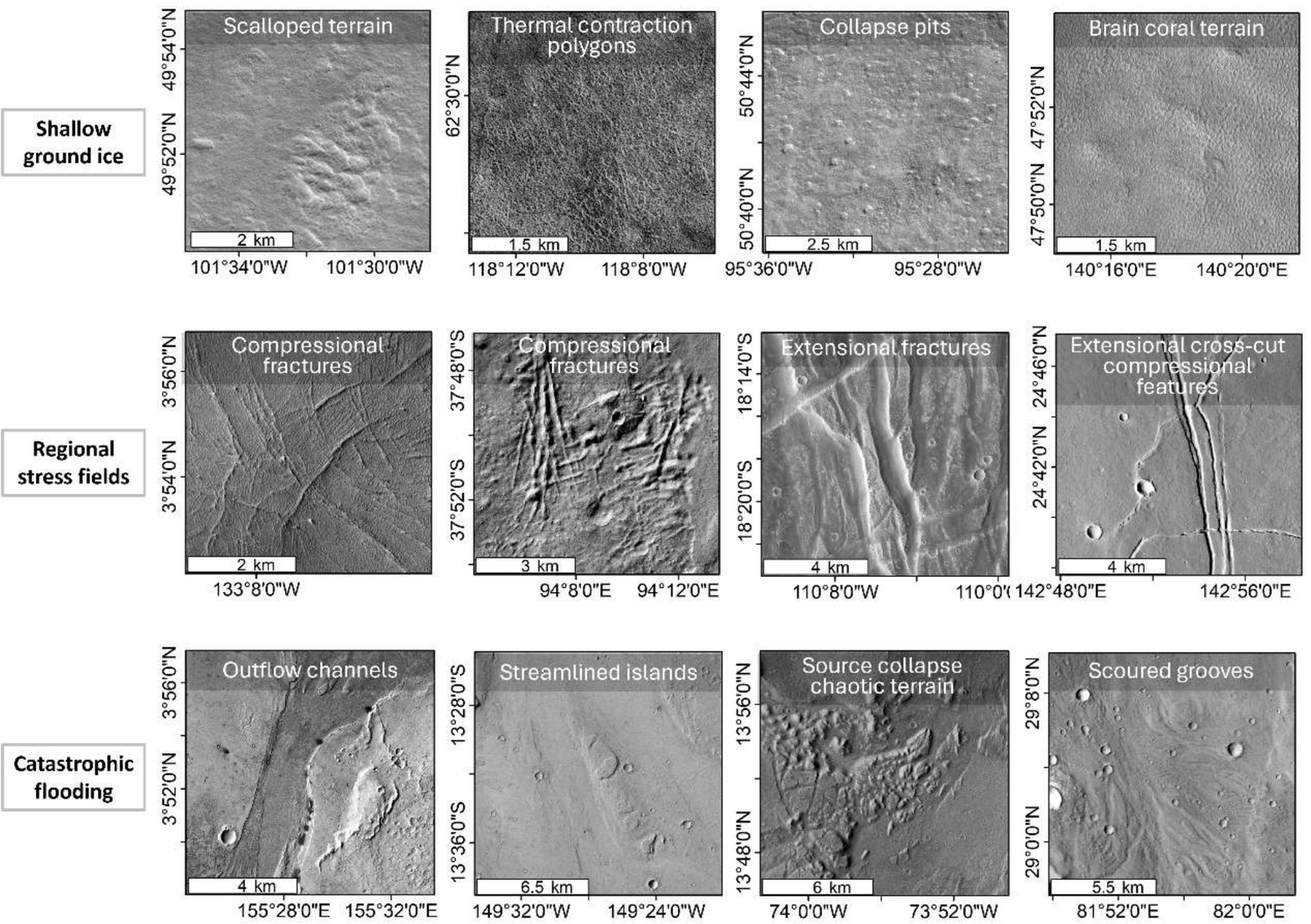}
\caption{\textbf{Process-based geomorphological mapping enabled
by MarScope.} Examples retrieved using formation-mechanism--oriented
queries, including shallow ground ice, regional stress fields, and
catastrophic flooding.}
\label{fig:4}
\end{figure}

This process-based mapping allows hypotheses about surface processes to be explored directly at planetary scale. 
As shown in Figure \ref{fig:4}, single process-oriented queries retrieve morphologically diverse yet genetically related features: ice-related queries return multiple ice-associated and collapse landforms; stress-related queries identify coherent networks of compressional and extensional structures; and flow-related queries retrieve channels, streamlined forms, and
source-collapse terrains across distinct geological settings. 
Such integrative retrievals make it possible to examine surface processes coherently across morphologically distinct landforms at planetary scale, enabling analyses that were previously impractical.

\paragraph*{Mapping of rare and previously unnamed landforms through visual search.}
Beyond its capability for large-scale mapping, MarScope enables the
mapping of previously understudied or unclassified landforms. Through
image-to-image retrieval, the system searches the Martian surface by
visual similarity rather than predefined labels, allowing recognition of
features rarely documented in existing catalogues.

\begin{figure}[htbp]
\centering
\includegraphics[width=0.9\linewidth]{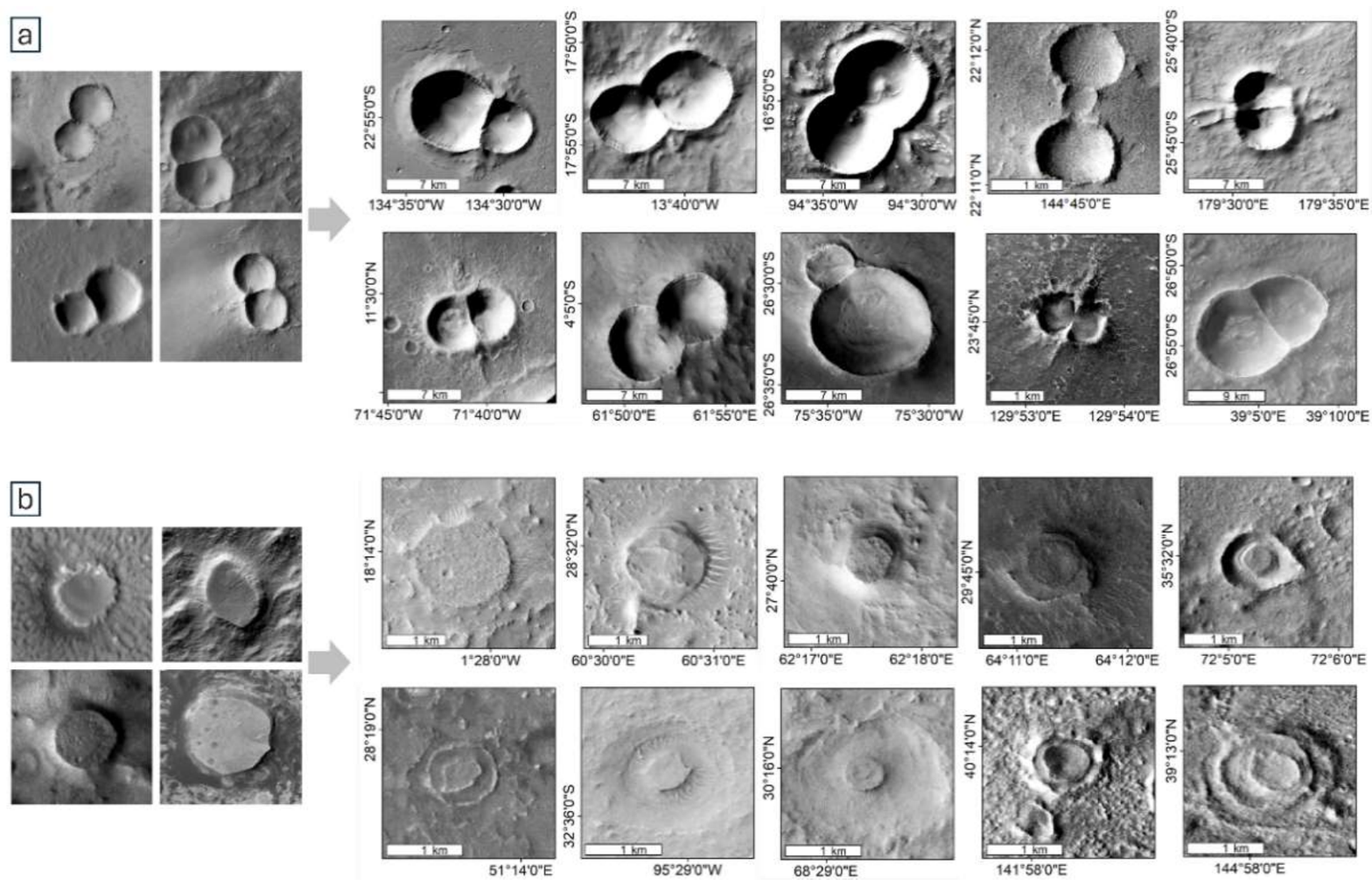}
\caption{\textbf{Visual discovery of rare Martian landforms
using MarScope.} Retrieval of (\textbf{A}) doublet craters and (\textbf{B}) inverted
craters, showing reference examples (left) and morphologically similar
features identified across global CTX data (right).}
\label{fig:5}
\end{figure}

A representative example is the detection of doublet craters (Figure \ref{fig:5}A),
paired impact structures formed by the near-simultaneous collision of
binary asteroids \cite{vavilov2022evidence}. Recognition of them provides
insight into the population and orbital dynamics of binary impactors and
the mechanics of double impacts under Martian conditions. Using verified
examples as visual references, MarScope retrieves morphologically
similar crater pairs across global datasets within seconds, enabling the
construction of a statistically robust inventory of doublet impacts.
Another example is the inverted crater (Figure \ref{fig:5}B), where sedimentary
infill became indurated and more resistant to erosion than the
surrounding terrain \cite{pain2007inversion}. As deflation stripped away
softer materials, the hardened fill remained as a raised landform.
Through visual similarity search, MarScope identifies analogous
structures, offering new means to study sedimentary infill, diagenesis,
and erosional modification on Mars.

These examples show that visual similarity search allows planetary
surfaces to be explored beyond predefined taxonomies, making it possible
to identify geomorphological features that are difficult to capture with
conventional, label-driven mapping.

\subsection*{Discussion}

This work moves planetary mapping beyond predefined, morphology-centric
schemes toward open-ended, language-guided mapping. By enabling natural
language to serve as an interface to planetary-scale image archives,
MarScope facilitates near-instantaneous generation of global
geomorphological distribution maps, allowing rapid exploration of
planetary surfaces. Through formation mechanisms search, MarScope
enables the mapping of features based on their inferred formation
processes, rather than predefined shapes or labels. Additionally, the
platform supports the identification of rare, transitional, and
previously understudied features, enabling the exploration of landforms
that may have been overlooked in traditional mapping efforts. This
formulation fosters integrative exploration of planetary surfaces
without requiring explicit labels or task-specific training, opening new
pathways for scientific discovery.

As a foundation model, MarScope can serve as the backbone for
specialized models, fine-tuned to address specific mission goals or
scientific inquiries, thereby providing a versatile infrastructure that
reduces the need for extensive custom model development for each new
planetary dataset. This adaptability means that MarScope can support a
broad spectrum of research, from basic planetary mapping to more complex
tasks, such as studying planetary habitability, resource distribution,
and surface dynamics over time.

MarScope's design focuses on broad, open-ended discovery rather than
highly precise detection of specific landforms, which makes it less
suited for tasks requiring pixel-level accuracy. The system identifies
the presence of features within image tiles but does not provide
pixel-level delineation. Furthermore, its fixed tile sizes (0.2° and
0.02°) reflect a trade-off between global coverage and local detail,
which may result in the underrepresentation of features that are either
larger than a tile or smaller than the effective resolution. These
limitations are inherent in the framework's general-purpose design,
which is optimized for flexible, large-scale exploration rather than
specialized feature extraction.

MarScope's performance is shaped by the quality and diversity of its
training data, including image resolution, spatial coverage, and
linguistic descriptions of surface features. Ongoing efforts to expand
the training corpus with higher-resolution imagery, more diverse feature
descriptions, and human-in-the-loop validation are expected to enhance
robustness, interpretability, and retrieval fidelity. These improvements
will further solidify MarScope's role as an exploratory interface,
complementing, rather than replacing, detailed physical modeling or
expert interpretation.

Although demonstrated here using Mars as a case study, MarScope's
underlying approach represents a paradigm shift in how machine
intelligence can mediate access to vast geospatial archives. This
paradigm enables scalable, hypothesis-driven exploration across
planetary bodies and scientific domains, where mapping relies on
semantic reasoning rather than rigid classification. Beyond planetary
science, this approach holds promise for Earth observation and other
fields, where large-scale, heterogeneous datasets require high-level
conceptual navigation.

\subsection*{Methods}

\paragraph*{Construction of planetary vision-language platform training dataset.}
This study developed a planetary-scale multimodal dataset to train
MarScope. Although MarScope is designed for Martian surface analysis,
its training data are not limited to Mars. To preserve cross-planet
extensibility, this study constructed a planetary-scale multimodal
dataset comprising more than 200,000 image-text pairs that capture
geomorphological features from Mars, the Moon, Mercury, and several icy
satellites. Images were collected from high-resolution orbital datasets
including HiRISE website, peer-reviewed literature and official NASA and
ESA releases.

Besides, a large language model (LLM)-assisted curation workflow was
employed to address the redundant and inconsistent information in the
original texts. The LLM performed relevance filtering, key-phrase
extraction, and semantic rewriting to refine each
caption \cite{khattak2024unimed}. Additional data
augmentation \cite{khattak2024unimed}, through synonym expansion and
paraphrasing, was used to enhance linguistic diversity and reduce
overfitting to specific wording. The resulting dataset spans a broad
range of geomorphological classes, imaging geometries, and illumination
conditions, enabling the model to generalize across planetary
environments and spatial scales.

\paragraph*{Vision-language platform semantic encoding and feature alignment.}
MarScope was trained with a contrastive learning objective that aligns
visual and textual embeddings within a shared semantic
space \cite{radford2021learning_arxiv}. The network maximizes similarity between
matched image--text pairs and minimizes it between mismatched pairs.
Each planetary image patch is then represented as a high-dimensional
embedding vector.

A zero-shot classification setup was used to identify the optimal model
checkpoint \cite{radford2021learning}. In this setting, the pretrained encoder
was applied directly without additional fine-tuning, and image
embeddings were compared against textual prompts representing
geomorphological classes (e.g., ``crater,'' ``yardang,'' ``transverse
aeolian ridge'') from benchmark Mars datasets such as
DoMars16k \cite{wilhelm2020domars16k} and the HiRISE v3
dataset \cite{wagstaff2021mars}. Measures of classification accuracy and the
semantic alignment between visual and textual representations were used
solely to determine the checkpoint with the most reliable parameter
configuration for deployment.

\paragraph*{Platform implementation: hierarchical search framework and near-instantaneous query system.}
To enable large-scale semantic retrieval and analysis, all image
embeddings produced by the trained model were indexed within a Facebook
AI Similarity Search (FAISS)-based approximate nearest-neighbor (ANN)
database \cite{douze2025faiss}. The global Mars CTX
mosaic \cite{malin2007context, dickson2023release} was divided into overlapping tiles
(\textasciitilde130 Million) at two hierarchical resolutions (0.2° and
0.02°) to balance global coverage with local detail. Each tile's visual
embedding, produced by the trained encoder, was stored as a
high-dimensional vector linked to geospatial metadata. This approach
effectively reduced the raw imagery by approximately 160×, from 900 GB
to 5.7 GB at 0.2° resolution and 2.2 TB to 14 GB at 0.02°, while
preserving key semantic information.

Cross-modal retrieval was implemented within the shared embedding space,
where similarity between image and text embeddings is measured using
cosine distance. Text queries were first encoded using the trained text
encoder, and the resulting vectors were used to retrieve the most
similar image embeddings \cite{gu2024language}. This procedure underpins
all retrieval modes \cite{gu2024language}, including image, text, and
image-text multimodal searches.

To examine retrieval consistency and refine the similarity threshold,
the top ten matches and ten near-threshold samples were visually
inspected. The near-threshold cases were used to adjust the decision
boundary and ensure stable retrieval behavior across heterogeneous
landform types.

For visualization and subsequent geospatial analysis, retrieved
embeddings were mapped back to their original planetary coordinates to
generate spatial distribution map and density map. In addition, all
retrieval results, including coordinates information and image clips,
can be exported for downstream analyses.

\paragraph*{Evaluation with published catalogue.}
To quantitatively evaluate the retrieval performance of MarScope, we
compare our retrieval results with previously published global
catalogues \cite{morgan2022global, souness2012inventory, roback2021controls, mills2024global, liu2020mapping, bickel2025streaks}. Six types were selected for
validation: alluvial fans \cite{morgan2022global}, glacier-like
forms \cite{souness2012inventory}, landslides \cite{roback2021controls}, pitted
cones \cite{mills2024global}, yardangs \cite{liu2020mapping} and dark slope
streak \cite{bickel2025streaks}. For each landform type, the published
catalogue provides a set of reference points
$G = \{ g_{1},g_{2},...,g_{M}\}$, which is expressed as planetocentric
latitude and longitude. Since every image patch in our MarScope database
is georeferenced, we transfer the top-$K$ retrieved image patches
to a set of center coordinates $P = \{ p_{1},p_{2},...,p_{M}\}$.
Consequently, we formulate our evaluation as a point-set matching
problem between the predicted locations and the ground truth catalog.

Given the continuous nature of geospatial coordinates, it is infeasible
to exact positional coincidence. Instead, we develop a proximity-based
matching criterion with a tolerance radius $r$ (set to 0.5° in this
study) during matching. In other words, we regard a retrieved patch as
correctly matched if the nearest-neighbor distance to the corresponding
ground-truth point is below the threshold. According to the above
criterion, we define the following three metrics:

\begin{equation}
Precision@K = \frac{1}{K}\sum_{i = 1}^{K}\mathbb{I}\left( \min_{g \in G} \parallel p_{i} - g \parallel \leq r \right)
\end{equation}

\begin{equation}
Recall@K = \frac{1}{M}\sum_{j = 1}^{M}\mathbb{I}\left( \min_{p \in P_{K}} \parallel p - g_{j} \parallel \leq r \right)
\end{equation}

\begin{equation}
F1@K = 2 \cdot \frac{Precision@K \cdot Recall@K}{Precision@K + Recall@K}
\end{equation}

where $\mathbb{I(} \cdot )$ denotes the indicator function. Here,
precision serves as a measure of retrieval fidelity, which indicates the
percentage of candidates spatially validated by the catalogue.
Conversely, Recall quantifies the completeness of the global mapping,
representing the fraction of catalogue entries that are successfully
recovered by MarScope. Since increasing the retrieval depth $K$
inherently presents a trade-off between Recall and Precision, we utilize
the F1@K to identify the optimal sensitivity.


\section*{Acknowledgments}
We thank the mission teams of NASA for providing open-access planetary datasets used in this study, including HiRISE, CTX and other publicly available archives. Language polishing and refinement of the manuscript were assisted by ChatGPT (version 5.2), and the authors take full responsibility for the content.

\paragraph*{Funding:}
This work was funded by the National Natural Science Foundation of China (42578017, 42272274, 42273041) and Key Technology Research Project of TW-3 (TW3004).

\paragraph*{Competing interests:}
The authors declare no competing interests.

\paragraph*{Data and materials availability:}
The MarScope can be researched at \url{http://marscope.site/}.
Remote sensing datasets (CTX imagery) are publicly available at \url{https://murray-lab.caltech.edu/CTX/}.
\bibliographystyle{unsrt} 
\bibliography{references}

\end{document}